\documentclass{article}

\usepackage[final]{nips_2016}

\usepackage[utf8]{inputenc} 
\usepackage[T1]{fontenc}    
\usepackage{hyperref}       
\usepackage{url}            
\usepackage{booktabs}       
\usepackage{amsfonts}       
\usepackage{nicefrac}       
\usepackage{microtype}      
\usepackage[usenames,dvipsnames]{color}
\usepackage{graphicx}
\usepackage{subcaption}
\usepackage[group-separator={,}]{siunitx}
\usepackage{amsmath}
\usepackage{amssymb}
\usepackage{algorithm}
\usepackage{algpseudocode}

\newif\ifcomments
\commentsfalse

\ifcomments
\newcommand{\jmnote}[1]{\textcolor{Green}{\textbf{[JM: #1]}}}
\newcommand{\stnote}[1]{\textcolor{blue}{\textbf{ST: #1}}}
\else
\newcommand{\jmnote}[1]{\textcolor{Green}{}}
\newcommand{\stnote}[1]{\textcolor{blue}{}}
\fi

\DeclareMathOperator*{\argmax}{arg\,max} 

\title{Implementing the Deep Q-Network}

\author{
  Melrose Roderick \\
  Humans To Robots Laboratory\\
  Brown University\\
  Providence, RI 02912 \\
  \texttt{melrose\_roderick@brown.edu} \\
  \And
  James MacGlashan \\
  Humans To Robots Laboratory\\
  Brown University\\
  Providence, RI 02912 \\
  \texttt{jmacglashan@gmail.com} \\
  \And
  Stefanie Tellex \\
  Humans To Robots Laboratory\\
  Brown University\\
  Providence, RI 02912 \\
  \texttt{stefie10@cs.brown.edu} \\
}

\begin{document}

\maketitle

\begin{abstract}
  The Deep Q-Network proposed by \citet{mnih2015human} has become a benchmark and building point for much deep reinforcement learning research.
  However, replicating results for complex systems is often challenging since original scientific publications are not always able to describe in detail every important parameter setting and software engineering solution.
  In this paper, we present results from our work reproducing the results of the DQN paper.
  We highlight key areas in the implementation that were not covered in great detail in the original paper to make it easier for researchers to replicate these results, including termination conditions and gradient descent algorithms.
  Finally, we discuss methods for improving the computational performance and provide our own implementation that is designed to work with a range of domains, and not just the original Arcade Learning Environment~\citep{bellemare13arcade}.
\end{abstract}

\section{Introduction}

Over the past few years, deep reinforcement learning has gained much popularity as it has been shown to perform better than previous methods on domains with very large state-spaces.
In one of the earliest deep reinforcement learning papers (hereafter the DQN paper), \citet{mnih2015human} presented a method for learning to play Atari 2600 video games, using the Arcade Learning Environment (ALE)~\citep{bellemare13arcade}, from image and performance data alone using the same deep neural network architecture and hyper-parameters for all the games.
DQN outperformed previous reinforcement learning methods on nearly all of the games and recorded better than human performance on most.
As many researchers tackle reinforcement learning problems with deep reinforcement learning methods and propose alternative algorithms, the results of the DQN paper are often used as a benchmark to show improvement.
Thus, implementing the DQN algorithm is important for both replicating the results of the DQN paper for comparison and also building off the original algorithm.
One of the main contributions of the DQN paper was finding ways to improve stability in their artificial neural networks during training.
There are, however, a number of other areas in the implementation of this method that are crucial to its success, which were only mentioned briefly in the paper.

We implemented a Deep Q-Network (DQN) to play the Atari games and replicated the results of \citet{mnih2015human}.
Our implementation, available freely online,\footnote{\url{www.github.com/h2r/burlap_caffe}} runs around 4x faster than the original implementation.
Our implementation is also designed to be flexible to different neural net network architectures and problem domains outside of ALE.
In replicating these results, we found a few key insights into the process of implementing such a system.
In this paper, we highlight key techniques that are essential for good performance and replicating the results of \citet{mnih2015human}, including termination conditions and gradient descent optimization algorithms, as well as expected results of the algorithm, namely the fluctuating performance of the network.

\section{Related Work}{}

The Markov Decision Process (MDP) \citep{bellman1957markovian} is the typical formulation used for reinforcement learning problems.
An MDP is defined by a five-tuple $(\mathcal{S, A, T, R, E})$;
$\mathcal{S}$ is the agent's state-space;
$\mathcal{A}$ is the agent's action-space;
$\mathcal{T}(s, a, s')$ represents the transition dynamics, which returns the probability that taking action $a$ in state $s$ will result in the state $s'$;
$\mathcal{R}(s, a, s')$ is the reward function, which returns the reward received when transitioning to state $s'$ after taking action $a$ in state $s$;
and $\mathcal{E} \subset \mathcal{S}$ is the set of terminal states, which once reached prevent any future action or reward.
The goal of planning in an MDP is to find a policy $\pi : S \rightarrow A$, a mapping from states to actions, that maximizes the expected future discounted reward when the agent chooses actions according to $\pi$ in the environment. A policy that maximizes the expected future discounted reward is an optimal policy and is denoted by $\pi^*$.

A key concept related to MDPs is the Q-function, $Q^\pi : S \times A \rightarrow \mathbb{R}$, that defines the expected future discounted reward for taking action $a$ in state $s$ and then following policy $\pi$ thereafter. According to the Bellman equation, the Q-function for the optimal policy (denoted $Q^*$) can be recursively expressed as:
\begin{equation}
Q^*(s, a) = \sum_{s' \in S} T(s, a, s') \left [ R(s, a, s') + \gamma \max_{a'} Q^*(s', a') \right ]
\end{equation}
where $0 \leq \gamma \leq 1$ is the discount factor that defines how valuable near-term rewards are compared to long-term rewards. 
Given $Q^*$, the optimal policy, $\pi^*$, can be trivially recovered by greedily selecting the action in the current state with the highest Q-value: $\pi^*(s) = \argmax_a Q^*(s, a)$. This property has led to a variety of learning algorithms that seek to directly estimate $Q^*$, and recover the optimal policy from it. Of particular note is Q-Learning~\citep{watkins1989learning}. 

In Q-Learning, an agent begins with an arbitrary estimate ($Q_0$) of $Q^*$ and iteratively improves its estimate by taking arbitrary actions in the environment, observing the reward and next state, and updating its Q-function estimate according to
\begin{equation}
Q_{t+1}(s_t, a_t) \gets Q_t(s_t, a_t) + \alpha_t \left[ r_{t+1} + \gamma \max_{a'} Q_t(s_{t+1}, a') - Q_t(s_t, a_t) \right],
\end{equation}
where $s_t$, $a_t$, $r_t$ are the state, action, and reward at time step $t$, and $\alpha_t \in (0, 1]$ is a step size smoothing parameter.
Q-Learning is guaranteed to converge to $Q^*$ under the following conditions: the Q-function estimate is represented tabularly (that is, a value is associated with each unique state-action pair), the agent visits each state and action infinitely often, and $\alpha_t \rightarrow 0$ as $t \rightarrow \infty$.
When the state-space of a problem is large (or infinite), Q-learning's $Q^*$ estimate is often implemented with function approximation, rather than a tabular function, which allows generalization of experience.
However, estimation errors in the function approximation can cause Q-learning, and other ``off policy'' methods, to diverge~\citep{baird1995residual}, requiring careful use of function approximation.

\section{Deep Q-Learning}

\begin{algorithm}[t]
\begin{algorithmic}
\State Initialize replay memory $D$ to capacity $N$
\State Initialize action-value function $Q$ with random weights $\theta$
\State Initialize target action-value function $\hat Q$ with weights $\theta^{-} = \theta$
\For{episode 1, $M$}
Initialize sequence $s_1 = \{ x_1 \}$ and preprocessed sequence $\phi_1 = \phi(s_1)$
\For{$t = 1, T$}
\State With probability $\varepsilon$ select a random action $a_t$
\State otherwise select $a_t = \argmax_a Q(\phi(s_t), a; \theta)$
\State Execute action $a_t$ in the emulator and observe reward $r_t$ and image $x_{t+1}$
\State Set $s_{t+1} = s_t, a_t, x_{t+1}$ and preprocess $\phi_{t+1} = \phi(s_{t+1})$
\State Store experience $(\phi_t, a_t, r_t, \phi_{t+1})$ in $D$
\State Sample random minibatch of experiences $(\phi_j, a_j, r_j, \phi_{j+1})$ from $D$
\State Set $y_j = \begin{cases}
r_j & \text{if episode terminates at step $j+1$}\\
r_j + \gamma \max_{a'} \hat Q(\phi_{j+1}, a'; \theta^{-}) & \text{otherwise}
\end{cases}$
\State Perform a gradient descent step on $(y_j - Q(\phi_j, a_j ; \theta))^2$ with respect to the weights $\theta$
\State Every $C$ steps reset $\hat Q = Q$
\EndFor
\EndFor
\end{algorithmic}
\caption{Deep Q-learning with experience replay}
\label{alg:dqn}
\end{algorithm}

Deep Q-Learning (DQN)~\citep{mnih2015human} is a variation of the classic Q-Learning algorithm with 3 primary contributions: (1) a deep convolutional neural net architecture for Q-function approximation; (2) using mini-batches of random training data rather than single-step updates on the last experience; and (3) using older network parameters to estimate the Q-values of the next state.
Pseudocode for DQN, copied from \citet{mnih2015human}, is shown in Algorithm~\ref{alg:dqn}.
The deep convolutional architecture provides a general purpose mechanism to estimate Q-function values from a short history of image frames (in particular, the last 4 frames of experience). The latter two contributions concern how to keep the iterative Q-function estimation stable. 

In supervised deep-learning work, performing gradient descent on mini-batches of data is often used as a means to efficiently train the network. In DQN, it plays an additional role.
Specifically, DQN keeps a large history of the most recent experiences, where each experience is a five-tuple $(s, a, s', r, T)$, corresponding to an agent taking action $a$ in state $s$, arriving in state $s'$ and receiving reward $r$; and $T$ is a boolean indicating if $s'$ is a terminal state.
After each step in the environment, the agent adds the experience to its memory.
After some small number of steps (the DQN paper used 4), the agent randomly samples a mini-batch from its memory on which to perform its Q-function updates.
Reusing previous experiences in updating a Q-function is known as {\em experience replay}~\citep{lin1992self}.
However, while experience replay in RL was typically used to accelerate the backup of rewards, DQN's approach of taking fully random samples from its memory to use in mini-batch updates helps decorrelate the samples from the environment that otherwise can cause bias in the function approximation estimate.

The final major contribution is using older, or ``stale,'' network parameters when estimating the Q-value for the next state in an experience and only updating the stale network parameters on discrete many-step intervals. This approach is useful to DQN, because it provides a stable training target for the network function to fit, and gives it reasonable time (in number of training samples) to do so. Consequently, the errors in the estimation are better controlled.

Although these contributions and overall algorithm are straightforward conceptually, there are number of important details to achieving the same level of performance reported by \citet{mnih2015human}, as well as important properties of the learning process that a designer should keep in mind. We describe these details next.

\subsection{Implementation Details} 



Large systems, such as DQN, are often difficult to implement since original scientific publications are not always able to describe in detail every important parameter setting and software engineering solution.
Consequently, some important low-level details of the algorithm are not explicitly mentioned or fully clarified in the DQN paper.
Here we highlight some of these key additional implementation details, which are provided in the original DQN code.\footnote{\url{www.github.com/kuz/DeepMind-Atari-Deep-Q-Learner}}

Firstly, every episode is started with a random number of ``No-op'' low-level Atari actions (in contrast to the agent's actions which are repeated for $4$ frames) between $0$ and $30$ in order to offset which frames the agent sees, since the agent only sees every $4$ Atari frames.
Similarly, the $m$ frame history used as the input to the CNN is the last $m$ frames that the agent sees, not the last $m$ Atari frames.
Additionally, before any gradient descent steps, a random policy is run for $\num{50000}$ steps to fill in some experiences in order to avoid over-fitting to early experiences.

Another parameter worth noting is the network update frequency.
The original DQN implementation only chose to take a gradient descent step every $4$ environment steps of the algorithm as opposed to every step, as Algorithm \ref{alg:dqn} might suggest.
Not only does this greatly increase the training speed (since learning steps on the network are far more expensive than forward passes), it also causes the experience memory to more closely resemble the state distribution of the current policy (since 4 new frames are added to the memory between training steps as opposed to 1) and may prevent the network from over-fitting.  \stnote{Are there results for this?  Could there be?  (I know this is a lot of work and probably not worth the tiem but wanted to ask the question...)}


\subsection{The Fluctuating Performance of DQN}
A common belief for new users of DQN is that performance should fairly stably improve as more training time is given. Indeed, average Q-learning learning curves in tabular settings are typically fairly stable improvements and supervised deep-learning problems also tend have fairly steady average improvement as more data becomes available. However, it is not uncommon in DQN to have ``catastrophic forgetting'' in which the agent's performance can drastically drop after a period of learning. For example, in Breakout, the DQN agent may reach a point of averaging a high score over $400$, and then, after another large batch of learning, it might be averaging a score of only around $200$. The solution \citet{mnih2015human} propose to this problem is to simply save the network parameters that resulted in the best test performance.

One of the reasons this forgetting occurs is the inherent instability of approximating the Q-function over a large state-space using these Bellman updates.
One of the main contributions of \citet{mnih2015human} was fighting this instability using experience replay and stale network parameters, as mentioned above.
Additionally, \citet{mnih2015human} found that clipping the gradient of the error term to be between $-1.0$ and $1.0$ further improved the stability of the algorithm by not allowing any single mini-batch update to change the parameters drastically.
These additions, and others, to the DQN algorithm improve its stability significantly, but the network still experiences catastrophic forgetting.
\jmnote{I think I might suggest moving the gradient clipping out as its own unique thing to helping the stabilization. So whereas the main contributions to stabilization are the experience replay and stale network parameters, gradient clipping is also an unsung hero.}

Another reason this catastrophic forgetting occurs is that the algorithm is learning a proxy, the Q-values, for a policy instead of approximating the policy directly.
A side effect of this method of policy generation is a learning update could increase the accuracy of a Q-function approximator, while decreasing the performance of the resulting policy.
For example, say the true Q-value for some state, $s$, and actions, $a_1$ and $a_2$, are $Q^*(s, a_1) = 2$ and $Q^*(s, a_2) = 3$, so the optimal policy at state $s$ would be to choose action $a_2$.
Now say the Q-function approximator for these values using current parameters, $\theta$, estimates $\hat Q(s, a_1; \theta) = 0$ and $\hat Q(s, a_2; \theta) = 1$, so the policy chosen by this approximator will also be $a_2$
But, after some learning updates we arrive at a set of parameters $\theta'$, where $\hat Q(s, a_1; \theta) = 2$ and $\hat Q(s, a_2; \theta) = 1$.
These learning updates clearly decreased the error of the Q-function approximator, but now the agent will not choose the optimal action at state $s$. \jmnote{Doesn't the nature paper show something like the plots of error in Q-function over time showing that it's steadily improving? If so you may want to call that out as evidence that even though the Q-function improves, you might get a really bad result in performance.}

Furthermore Q-values for different actions of the same state can be very similar if any of these actions does not have a significant effect on near-term reward.
These small differences are the result of longer-term rewards and are therefore critical to the optimal policy.
The consequence of trying to learn an approximator for this type of function is that very small errors in the Q-values can result is very different policies, making it difficult to learn long-term policies.

As an example of this, we will consider Breakout.
Breakout is an Atari game where the player controls a paddle with the goal of bouncing a ball to destroy all the bricks on the screen without dropping the ball.
There is an optimal strategy which is to destroy the bricks on the side of the screen so that the ball can be bounced above the bricks.
When the ball is above the bricks, the Q-values are much higher than they are when the ball is below the bricks, so we would expect a policy which follows the true Q-values to quickly exploit this policy.
Every time your paddle bounces the ball, the direction does not affect short-term rewards as a brick will be broken every time you bounce the ball
But, the ball direction will affect the distant reward of bouncing the ball above the bricks by breaking the bricks on the side of the screen.
Thus, it is difficult for a Q-function approximator to learn this long-term optimal policy.

\begin{figure}
  \centering
  \captionsetup[subfigure]{labelformat=empty}
  \begin{subfigure}[b]{0.2\textwidth}
    \includegraphics[width=\textwidth]{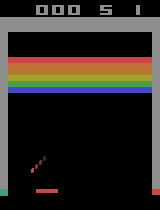}
  \end{subfigure}
  \begin{subfigure}[b]{0.35\textwidth}
    \includegraphics[width=\textwidth]{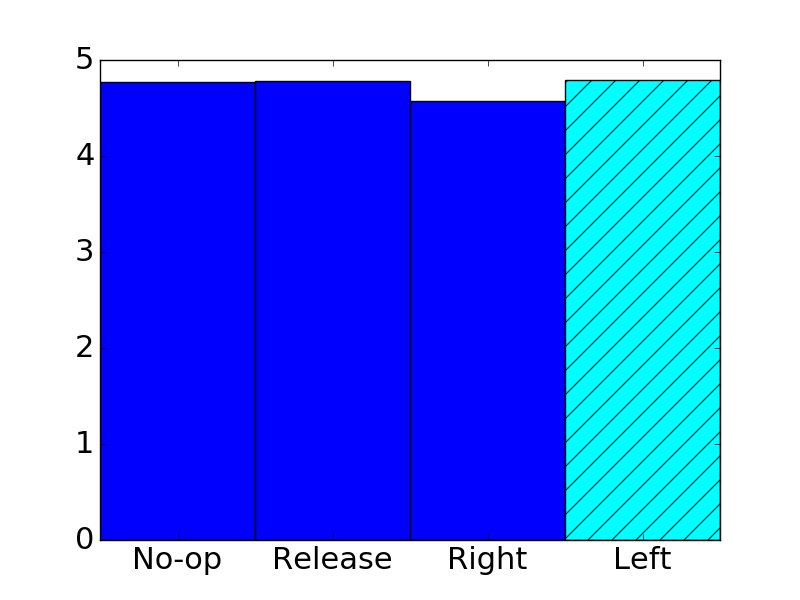}
  \end{subfigure}
  \begin{subfigure}[b]{0.35\textwidth}
    \includegraphics[width=\textwidth]{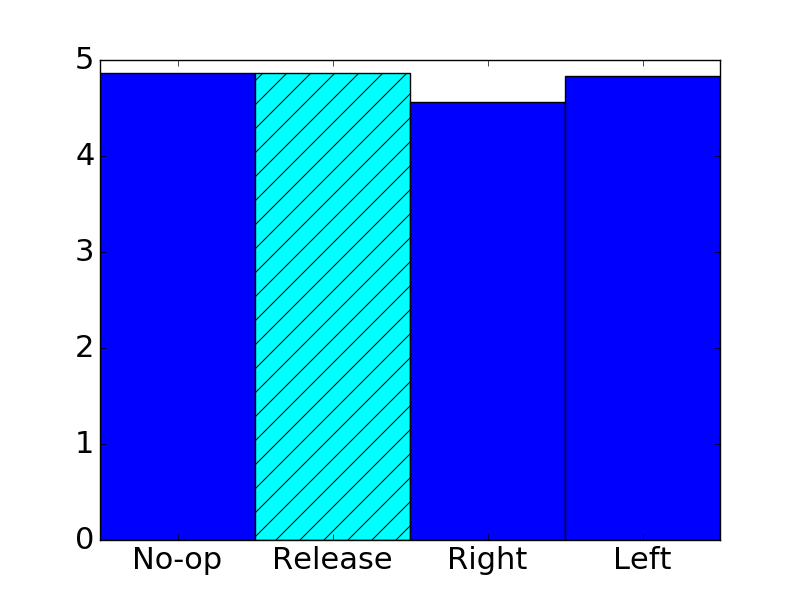}
  \end{subfigure}

  \begin{subfigure}[b]{0.2\textwidth}
    \includegraphics[width=\textwidth]{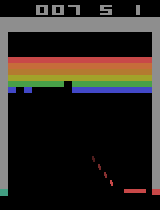}
    \caption{Example Frames}
  \end{subfigure}
  \begin{subfigure}[b]{0.35\textwidth}
    \includegraphics[width=\textwidth]{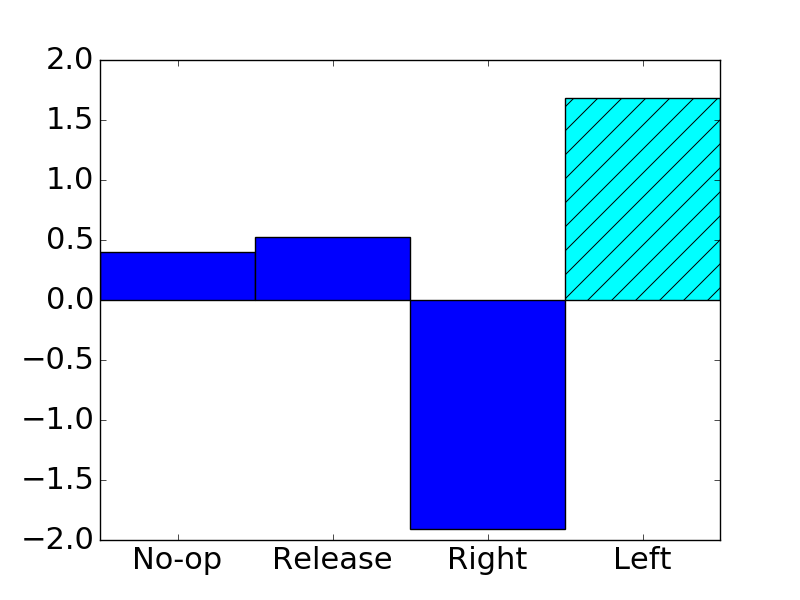}
    \caption{Best Network}
  \end{subfigure}
  \begin{subfigure}[b]{0.35\textwidth}
    \includegraphics[width=\textwidth]{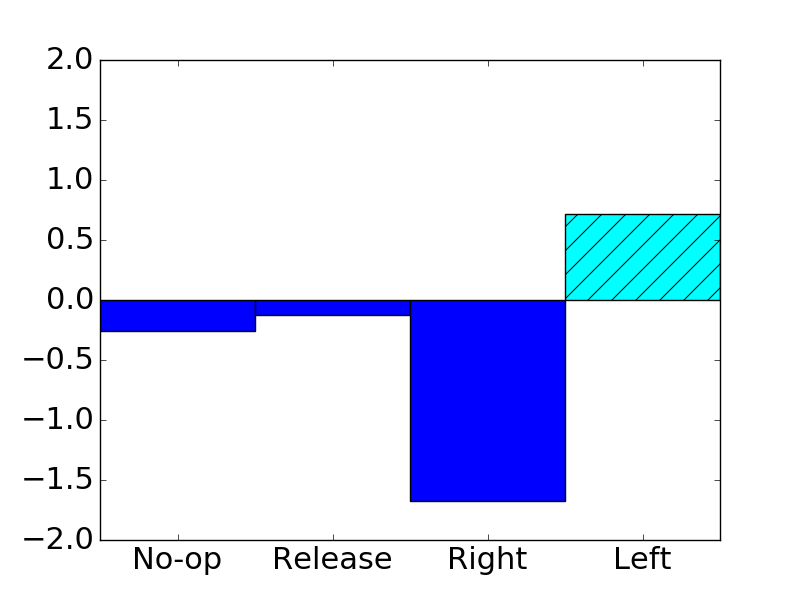}
    \caption{Worst Network}
  \end{subfigure}
  \caption{
  A comparison of calculated Q-values by the networks that received the best and worst performance during testing that had been trained for at least 30,000,000 steps.
  The lighter cross-hatched bar indicates the action with the highest Q-value.
  The top frame corresponds to a situation where the actions don't have a significant affect on the near-future reward, while the bottom one shows a situation where the left action must be made to not loose a life.
  The ``Release'' action releases the ball at the beginning of every round or does nothing (the same as ``No-op'') if the ball is already in play.
  }
  \label{fig:q_values}
\end{figure}

Figure \ref{fig:q_values} shows Q-values approximated by the best network and a network that performed poorly very late into training on the same inputs near the beginning of a Breakout game.
The first frame illustrates a scenario where any action could be made and the agent could still prevent the ball from falling a few actions into the future.
But the actions made before bouncing the ball also allow the agent to aim the ball.
The Q-values in this case are very similar for both networks, but the chosen actions are different.
In the second scenario, if the agent does not take the left action, the ball will be dropped, which is a terminal state.
In this case, the Q-values are much more distinct. 
Thus, this fluctuating performance is to be expected while running this algorithm.

\section{Machine Learning Libraries}

Our implementation uses the Brown-UMBC Reinforcement Learning and Planning (BURLAP) Java code library \citep{burlap}.
This library makes it easy to define a factored representation of an MDP and offers many well-known learning and planning algorithms as well as the machinery for creating new ones.

For running and interacting with the Atari video games, we used the Arcade Learning Environment (ALE) \citep{bellemare13arcade}.
ALE is a framework that provides a simple way to retrieve the screen and reward data from the Atari games as well as interact with the game through single actions.
We used ALE's FIFO Interface to interact ALE through Java.

To run and train our convolutional neural net, we used the Berkeley's Caffe (Convolutional Architecture for Fast Feature Embedding) library \citep{jia2014caffe}.
Caffe is a fast deep learning framework for constructing and training neural network architectures.
To interact with Caffe through our Java library, we used the JavaCPP library provided by Bytedeco.\footnote{\url{www.github.com/bytedeco/javacpp}}

\section{Results}

To measure our performance against that of \citet{mnih2015human}, we followed the same evaluation procedure as their paper on three games: Pong, Breakout, and Seaquest.
We trained the agent for $\num{50000000}$ steps (each step is 4 Atari frames) and tested performance every $\num{250000}$ steps.
We saved the network parameters that resulted in the best test performance.
We then evaluated the trained agent with the best performing network parameters on 30 games with and $\varepsilon$-greedy policy where $\varepsilon = 0.05$.
Each game was also initialized with a random number of ``No-op'' low-level Atari actions between $0$ and $30$.
We then took the average score of those games.

The comparison of our results and those of the DQN paper on Pong, Breakout, and Seaquest are shown in Table \ref{results}.
Each training process took about 3 days for our implementation and about 10 and a half days for the original implementation on our setup.

The differences in performance stem from the differences in gradient descent optimization algorithm and learning rate.
These differences are covered in more detail in section \ref{RMS}.

\begin{table}[t]
  \captionsetup{skip=8pt}
  \caption{Comparison of average game scores obtained by our DQN implementation and the original DQN paper.}
  \label{results}
  \centering
  \begin{tabular}{lll}
    \toprule
    Game     & Our implementation    & The original implementation \\
    \midrule
    Pong     & $19.7 \ (\pm 1.1)$    & $18.9 \ (\pm 1.3)$          \\
    Breakout & $339.3 \ (\pm 86.1)$  & $401.2 \ (\pm 26.9)$        \\
    Seaquest & $6309 \ (\pm 1027)$   & $5286 \ (\pm 1310)$         \\
    \bottomrule
  \end{tabular}
\end{table}

\section{Key Training Techniques}

While implementing our DQN, we found there were a couple methods that were only mentioned briefly in the DQN paper, but critical to the overall performance of the algorithm.
Here we present these methods and explain why they have such a strong impact on training the network.

\subsection{Termination on the Loss of Lives}

In most of the Atari games, there is a notion of ``lives'' for the player, which correspond to the number of times the player can fail (such as dropping the ball in Breakout or running into a shark in Seaquest) before the game is over.
To increase performance, \citet{mnih2015human} chose to count the loss of a life (in the games involving lives) as a terminal state in the MDP during training.
This termination condition was not mentioned in much detail in the DQN paper, but is essential for achieving their performance.

\begin{figure}
  \centering
  
  \begin{subfigure}[b]{0.48\textwidth}
    \includegraphics[width=\textwidth]{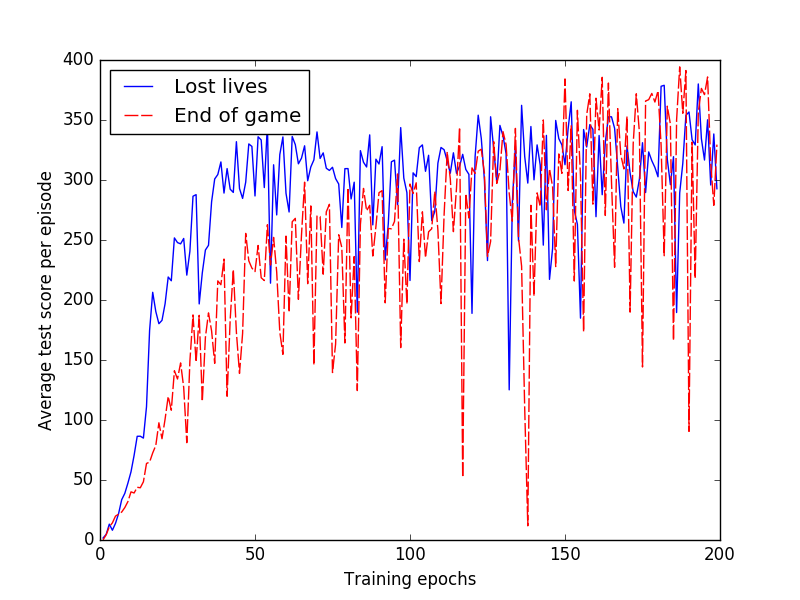}
    \caption{Breakout}
  \end{subfigure}
  \begin{subfigure}[b]{0.48\textwidth}
    \includegraphics[width=\textwidth]{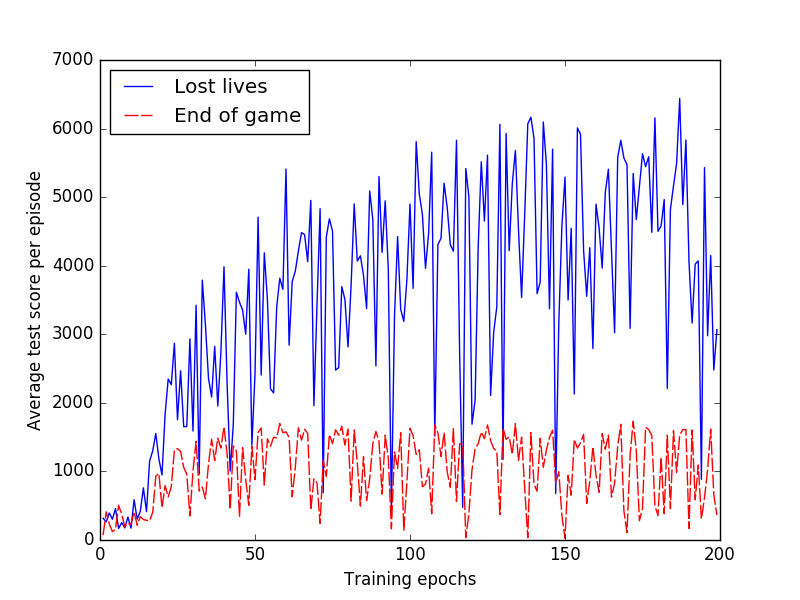}
    \caption{Seaquest}
  \end{subfigure}
  
  \caption{The average training test score received for Breakout and Seaquest at each test set when using lost lives as terminal states and when using the end of a game as terminal states (epoch = 250,000 steps).}
  \label{fig:lives-vs-no-lives}
\end{figure}

Figure \ref{fig:lives-vs-no-lives} illustrates the difference between training with and without counting losing lives as terminal states in both Breakout and Seaquest.
In Breakout, the average score of the learner that uses end of lives as terminal states increases much faster than the other learner.
However, around halfway through training, the other learner is achieving similar performance, but with much higher variance.
Seaquest is a much more complex game with many more moving sprites and longer episode lengths.
In this game the learner that uses lives as terminal states performs significantly better than the other learner throughout training.
These figures illustrate that this additional prior information greatly benefits early training and stability and, in the more complex games, significantly improves the overall performance.

A terminal state in an MDP, as mentioned above, signifies to the agent that no more reward can be obtained.
Almost all the Atari games give positive rewards (Pong is a notable exception where a reward of $-1$ is received when the enemy scores a point), and thus, this addition essentially informs the agent that losing a life should be avoided at all costs.  
This additional information given to the agent does seem reasonable: many human players know that loosing a life in an Atari game is bad the first time they play and it is difficult to imagine situations where the optimal policy would be to lose a life.

There are, however, a few theoretical issues with enforcing this
constraint.  The first being that this process is no longer Markovian
as the initial state distribution depends on the current policy.  An
example of this is in Breakout: If the agent performed well and broke many bricks before
losing a life, the new initial state for the next life will have many
fewer bricks remaining than if the agent performed poorly and broke very few bricks in the
previous life.  The other issue is that this signal gives strong
additional information to the DQN, making it challenging to extend to
domains where such strong signals are not available (e.g., real-world
robotics or more open-ended video games.)  

Although ALE stores the number of lives remaining for each game, it does not provide this information to all the interfaces.
To work around this limitation, we modified ALE's FIFO Interface to provide the number of lives remaining along with the screen, reward, and terminal state boolean.
Our fork that provides this data to the FIFO interface is available freely online.\footnote{\url{www.github.com/h2r/arcade-learning-environment}}

\subsection{Gradient Descent Optimization} \label{RMS}

One potential issue in using the hyper-parameters provided by \citet{mnih2015human} is that they are not using the same RMSProp definition that many deep learning libraries (such as Caffe) provide.
The RMSProp gradient descent optimization algorithm was originally proposed by Geoffrey Hinton.\footnote{\url{www.cs.toronto.edu/~tijmen/csc321/slides/lecture_slides_lec6.pdf}}
Hinton's RMSProp keeps a running average of the gradient with respect to each parameter.
The update rule for this running average can be written as:
\begin{equation}
MeanSquare(w, t) = \gamma \cdot MeanSquare(w, t-1) + (1 - \gamma) \cdot (\frac{\partial E}{\partial w}(t))^2 
\end{equation}
Here, $w$ corresponds to a single network parameter, $\gamma$ is the decay factor, and $E$ is the loss.
The parameters are then updated by:
\begin{equation}
w_t = w_{t-1} - \frac{\alpha}{\sqrt{MeanSquare(w, t) + \varepsilon}} \cdot \frac{\partial E}{\partial w}(t) 
\end{equation}
where $\alpha$ corresponds to the learning rate, and $\varepsilon$ is a small constant to avoid division by $0$.

Although \citet{mnih2015human} cite Hinton's RMSProp, they use a
slight variation on the algorithm.  The implementation of this can be
seen in their GitHub
repository\footnote{\url{www.github.com/kuz/DeepMind-Atari-Deep-Q-Learner}}
in the NeuralQLearner.lua file on lines 266-273.  \stnote{Nice!}  This variation adds
a momentum factor to the RMSProp algorithm that is updated as follows:
\begin{equation}
Momentum(w, t) = \eta \cdot Momentum(w, t-1) + (1 - \eta) \cdot \frac{\partial E}{\partial w}(t)
\end{equation}
Here, $\eta$ is the momentum decay factor.
The parameter update rule is then modified to:
\begin{equation}
w_t = w_{t-1} - \frac{\alpha}{\sqrt{MeanSquare(w, t) - (Momentum(w, t))^2 + \varepsilon}} \cdot \frac{\partial E}{\partial w}(t) 
\end{equation}

To account for this change in change in optimization algorithm, we had to modify the learning rate to something much lower than that of the \citet{mnih2015human} implementation (we used $0.00005$ as opposed to their $0.00025$).
We did not choose to implement this variant of RMSProp as it was not trivial to implement with the Java-Caffe bindings and Hinton's version produced similar results.

\section{Speed Performance}

Our implementation is training a bit less than 4x faster than the original implementation written in Lua using Torch.
The setup on which we tested these implementations is using two NVIDIA GTX 980 TI graphics cards along with an Intel i7 processor.
Our implementation runs at around $985$ Atari frames per second (fps) during training and $1584$fps during testing, while the Lua implementation runs at $271$fps during training and $586$fps during testing on our hardware (note that the algorithm only looks at every 4 frames, and so only 1 fourth of this number of frames are processed by the algorithm per second).
We attribute a large portion of this performance increase to cuDNN.

The NVIDIA CUDA Deep Neural Network library (cuDNN) is a proprietary NVIDIA library for running forward and backward passes on common neural network layers optimized specifically for the NVIDIA GPUs.
For both Torch and Caffe, cuDNN is available, but not used by default.
We compiled Caffe using cuDNN for our experiments, while the Lua implementation did not use this feature in Torch.
For comparison, when using Caffe without cuDNN, our implementation runs at around $268$fps during training and $485$fps during testing, which is a bit slower than the Lua implementation.

Another area that significantly increased the speed performance of our implementation was preallocating memory before training, which was also done in the original DQN implementation.
Allocating large amounts of memory is an expensive operation, so preallocating memory for large vectors, such as the experience memory and mini-batch data, and reusing it at each iteration significantly decreases this overhead.

\section{Conclusion}
In this paper we have presented a few key areas in the implementation of the DQN proposed by \citet{mnih2015human} that were essential to the overall performance of the algorithm, but were not covered in great detail in the original paper, in order to make it easier for researchers to implement their own versions of this algorithm.
We also highlighted some of the difficulties in approximating a Q-function with a CNN in such large state-spaces, namely the catastrophic forgetting.
We also have our implementation available freely online\footnote{\url{www.github.com/h2r/burlap_caffe}} and encourage researchers to use this as a tool for implementing novel algorithms as well as comparing performance with that of \citet{mnih2015human}.


\subsubsection*{Acknowledgments}
This material is based upon work supported by the National Science Foundation under grant numbers IIS-1637614 and IIS-1426452, and DARPA under grant numbers W911NF-10-2-0016 and D15AP00102.

\medskip

{
\bibliographystyle{plainnat}
\bibliography{reference}
}
\end{document}